%% file: iclr2024_conference.tex
\documentclass{article} 
\usepackage{iclr2024_conference,times}

\input{math_commands.tex}

\usepackage{hyperref}
\usepackage{url}

\usepackage{xspace}
\usepackage{nicefrac}
\usepackage{multirow}
\usepackage{makecell}
\usepackage{array}
\usepackage{graphicx}
\usepackage[capitalize,noabbrev]{cleveref}
\usepackage{subfigure}
\usepackage{authblk}
\PassOptionsToPackage{table}{xcolor}
\usepackage{xcolor} 
\usepackage{booktabs}

\newcommand{\methodname}{\emph{LazyLLM}\xspace}
\newcommand{\ttft}{\emph{TTFT}\xspace}
\newcommand{\prefilling}{\emph{prefilling}\xspace}
\newcommand{\decoding}{\emph{decoding}\xspace}

\makeatletter
\DeclareRobustCommand\onedot{\futurelet\@let@token\@onedot}
\def\@onedot{\ifx\@let@token.\else.\null\fi\xspace}

\def\eg{\emph{e.g}\onedot} 
\def\ie{\emph{i.e}\onedot} 
 \def\vs{\emph{vs}\onedot}
\def\wrt{w.r.t\onedot} 

\def\aka{\emph{a.k.a}\onedot}
\makeatother

\title{\methodname: Dynamic Token Pruning for Efficient Long Context LLM Inference}

\author[1]{Qichen Fu}
\author[1]{Minsik Cho}
\author[1]{Thomas Merth}
\author[1]{Sachin Mehta} 
\author[2\thanks{Work done while working at Apple.}]{\\Mohammad Rastegari}
\author[1]{Mahyar Najibi}

\affil[1]{Apple}
\affil[2]{Meta AI}

\iclrfinalcopy 
\begin{document}

\maketitle

\input{abstract}

\input{intro}

\input{related}

\input{method}

\input{experiments}

\input{conclusion}

\bibliography{citations}
\bibliographystyle{iclr2024_conference}

\appendix

\end{document}

%% file: math_commands.tex
\usepackage{amsmath,amsfonts,bm}

\def\eqref#1{equation~\ref{#1}}

\def\1{\bm{1}}

\def\vs{{\bm{s}}}

\DeclareMathAlphabet{\mathsfit}{\encodingdefault}{\sfdefault}{m}{sl}
\SetMathAlphabet{\mathsfit}{bold}{\encodingdefault}{\sfdefault}{bx}{n}



%% file: abstract.tex
\begin{abstract}

The inference of transformer-based large language models consists of two sequential stages: 1) a \prefilling stage to compute the KV cache of prompts and generate the first token, and 2) a \decoding stage to generate subsequent tokens.
For long prompts, the KV cache must be computed for all tokens during the \prefilling stage, which can significantly increase the time needed to generate the first token. Consequently, the \prefilling stage may become a bottleneck in the generation process. An open question remains whether all prompt tokens are essential for generating the first token. To answer this, we introduce a novel method, \methodname, that selectively computes the KV for tokens important for the next token prediction in both the \prefilling and \decoding stages. Contrary to static pruning approaches that prune the prompt at once, \methodname allows language models to dynamically select different subsets of tokens from the context in different generation steps, even though they might be pruned in previous steps. Extensive experiments on standard datasets across various tasks demonstrate that \methodname is a generic method that can be seamlessly integrated with existing language models to significantly accelerate the generation \textbf{without fine-tuning}. For instance, in the multi-document question-answering task, \methodname accelerates the \prefilling stage of the LLama 2 7B model by $2.34\times$ while maintaining accuracy.

\end{abstract}

%% file: intro.tex
\section{Introduction}

Standard prompt-based LLM inference has two sequential stages: \prefilling and \decoding, as shown in \cref{fig:generation_pipeline}. During the \prefilling stage, the model computes and saves the KV cache of each token from the prompt, and predicts the first token. We refer to the time taken during \prefilling stage as ``time-to-first-token'' (\ttft). Following the \prefilling stage is the \decoding stage, where the model reuses cached KVs to decode the next token iteratively until the stop criteria are met. 

During the \prefilling stage, all tokens from the prompt are used by all transformer layers. For long prompts, \ttft could be slow because state-of-the-art transformer-based LLMs are both deep and wide \citep{pope2023efficiently, kim2023full, aminabadi2022deepspeed}, and the cost of computing attention increases quadratically with the number of tokens in the prompts. For instance, Llama 2 \citep{touvron2023llama}, with 7 billion parameters, stacks 32 transformer layers with a model dimension of 4096. In this scenario, \ttft requires $21\times$ the walltime of each subsequent decoding step, and accounts for approximately 23\% of the total generation time on the LongBench benchmark\footnote{The average LongBench prompt length is $3376$ tokens and the average generation length is $68$ tokens.} \citep{bai2023longbench}.
Therefore, optimizing \ttft\ is a critical path toward efficient LLM inference \citep{nvidial40s}.

While optimizing LLM inference is an active area of research, many methods \citep{leviathan2023fast, cai2024medusa, zhang2024h2o, bhendawade2024speculative, li2024snapkv} have focused on improving inference speed during the \decoding stage. Yet, there is little attention given to improving \ttft. We note that some compression-based works implicitly improve the \ttft by reducing the size of LLMs \citep{frantar2022gptq, sun2023simple, ma2023llm}. However, an orthogonal line of research\citep{li2023compressing, jiang2023llmlingua, dao2022flashattention} investigates how \ttft can be improved given a static transformer architecture. Within this line of research, a natural question arises: Are all prompt tokens essential for generating the first token?

\input{figure_latex/generation_pipeline}

LLM profiling on the LongBench benchmark \citep{bai2023longbench} in \cref{fig:attention_sparsity} reveals that the attention scores of input tokens \wrt to the first generated token are very sparse, indicating that many tokens in the input prompt are redundant and can be removed without affecting the next token prediction. 
To this end, we propose \methodname, a novel, simple, yet effective technique tailored for speeding up \prefilling. As depicted in \cref{fig:teaser}, in each generation step, \methodname selectively computes the KV for tokens important for the next token prediction and ``lazily'' defers the computation of remaining tokens to later steps when they become relevant. 
We propose using the attention score of the prior transformer layer to measure the importance of tokens and progressively prune tokens along the depth of the transformer. 
In contrast to prompt compression works \citep{li2023compressing, jiang2023llmlingua, xu2023compress}, which permanently reduce the prompt for all the following generation steps, our method allows the model to revive previously pruned tokens, which we found crucial to retain accuracy. 
Extending progressive token pruning to all generation steps is non-trivial. Specifically, if a token is pruned at generation step $t$, and is revived at generation step $t' > t$, some hidden states would need to be recomputed during step $t'$.
To avoid such repetitive computation, we employ an additional caching mechanism, \emph{Aux Cache}, to cache the hidden states of pruned tokens. This enables a computationally efficient pathway to revive pruned tokens, and ensures that the worst runtime of \methodname\ is never slower than the baseline.

\input{figure_latex/attention_sparsity}

In summary, the advantages of \methodname are: (1) \textbf{Universal}: \methodname can be seamlessly integrated with any existing transformer-based LLM to improve inference speed, (2) \textbf{Training-free}: \methodname doesn't require any finetuning and can be directly integrated without any parameter modification, (3) \textbf{Effective}: Empirical results on 16 standard datasets across 6 different language tasks shows \methodname can improve the inference speed of the LLM during both \prefilling and \decoding stages.

%% file: figure_latex/generation_pipeline.tex
\begin{figure*}[t]
\centering
\includegraphics[width=\linewidth]{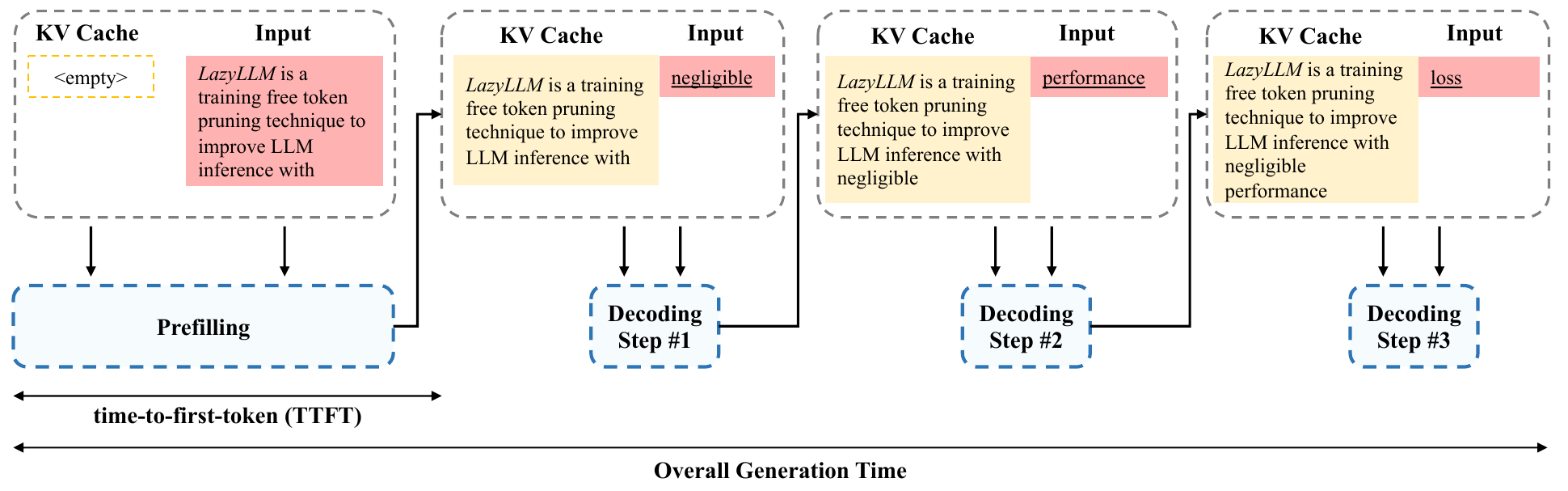}
\caption{Prompt-based LLM inference can be divided into two sequential stages: \prefilling and \decoding.
For long prompts, the first token generation during \prefilling stage could be slow.
As an example, for Llama 2 7B model \citep{touvron2023llama}, on average, the time to generate the first token requires $21\times$ the walltime of each subsequent decoding step and accounts for $23\%$ of the total generation time in the LongBench benchmark.
}
\label{fig:generation_pipeline}
\end{figure*} 

%% file: figure_latex/attention_sparsity.tex
\begin{figure*}[h]
\centering
\includegraphics[width=\linewidth]{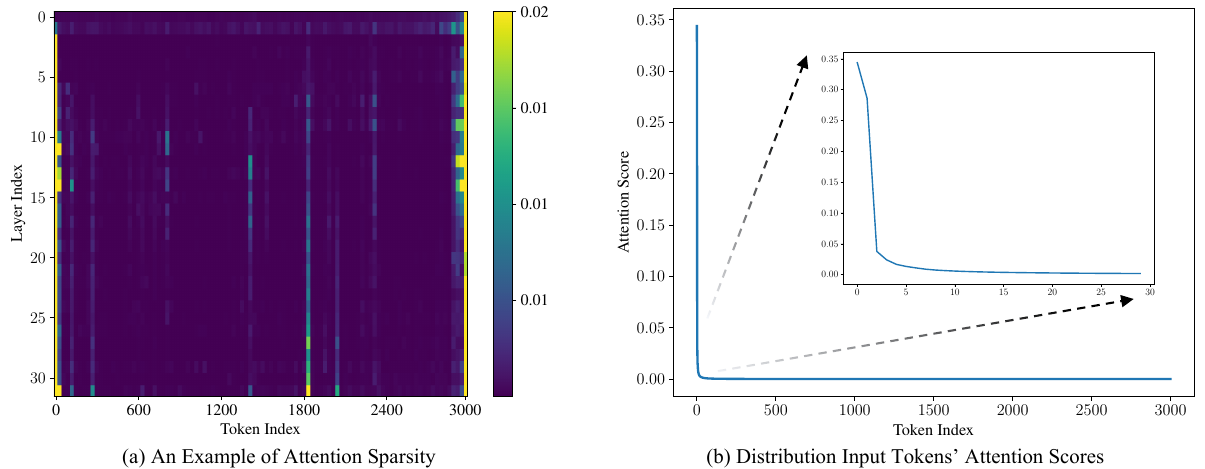}
\caption{We visualize the attention scores of input tokens in the prompt \wrt to the next token for each layer of Llama 2 7B\cite{touvron2023llama}. We also plot the distribution of the average attention score across all transformer layers. Result reveals that the attention scores of input tokens \wrt to the next token are very sparse, indicating that many tokens in the
input prompt are redundant and can be safely removed without affecting the next token prediction.}
\label{fig:attention_sparsity}
\end{figure*} 

%% file: related.tex
\section{Related Work}

The increase in the scale of large language models (LLMs) has greatly enhanced their performance but also introduced challenges with respect to their inference efficiency. The inference of generative LLMs consists of two distinct stages as depicted in \cref{fig:generation_pipeline}.
In particular, extensive computation is needed under long context scenarios to calculate the full KV cache during the \prefilling\ stage, resulting in a long time-to-first-token (\ttft). This delay causes users to wait several seconds after submitting a prompt before receiving any response from the agent, leading to a poor user experience.

\textbf{Efficient Long Context Inference.} Extensive work \citep{Merth2024SuperpositionPI, chen2023longlora, beltagy2020longformer, kitaev2020reformer} has been proposed to improve inference efficiency for long context applications by reducing the memory footprint and total computations. Some works have focused on tailoring the architecture of the transformer for long context input. For instance, \citep{beltagy2020longformer} introduces a drop-in replacement for standard self-attention and combines local windowed attention with task-motivated global attention. In parallel, Reformer \citep{kitaev2020reformer} replaces dot-product attention by one that uses locality-sensitive hashing to reduce its computational complexity. Though the above methods can speed up long context inference, they require significant model architecture change and re-training. This drawback makes them impractical to be applied to existing pre-trained LLMs.
Closer to our work are efficient techniques that optimize the KV cache \citep{zhang2024h2o, li2024snapkv, anagnostidis2024dynamic, nawrot2024dynamic} by minimizing the KV cache size and data transfer. However, these works only focus on accelerating decoding steps, which are not applicable to reducing \ttft.

\textbf{Token Pruning.} Previous studies on the sentence classification task \citep{kim2022learned, anagnostidis2024dynamic, he2021magic} has shown that not all tokens (\ie words) in an input sequence are necessary to make a successful prediction. This provides several possibilities for token pruning, which minimizes computational demands by selectively removing less important tokens during inference.
For example, \citep{kim2022learned} presents Learned Token Pruning which adaptively removes unimportant tokens as an input sequence passes through transformer layers. In parallel, \citep{he2021magic} proposes to reduce width-wise computation via token pruning for transformer-based models such as BERT~\citep{devlin2018bert}. These aforementioned approaches were designed for tasks requiring only a single iteration of processing, such as text classification. 
In this work, we extend the idea of token pruning to generative LLMs. Specifically, our method allows the model to dynamically choose different sets of tokens at each generation step, which is crucial to retaining the performance. 
Furthermore, we also introduce \emph{Aux Cache} to ensure that each token is computed
at most once along the whole generation, and ensure the worst runtime of our method is not slower than the baseline.

%% file: method.tex
\section{\methodname}

\input{figure_latex/teaser}

\subsection{Background on LLM Inference}
Generative LLM inference consists of two  stages: \prefilling and \decoding (see \cref{fig:generation_pipeline}). 
In the \prefilling stage, the model receives the prompt (a sequence of tokens) $\mathcal{T}=\{t_i\}_{i=1}^N$ of length N, where $t_i$ denotes a token and $N$ denotes the length of the prompt, then computes and saves the KV cache of each token, and produces the first token $t_{n+1}$. The transformer architecture commonly used in LLMs is a stack of layers where each layer shares the same architecture with a multiple-head self-attention mechanism followed by a multi-layer perception (MLP). 
The time of \prefilling is referred to as time-to-first-token (\aka \ttft).
Following the \prefilling is the \decoding steps, where the model appends the generated token $t_{n+1}$ to the input, and subsequently decodes the following token. The \decoding step is repeatedly performed until the stop criteria are met. While the formula of each decoding step is similar to \prefilling, the amount of its computation is significantly lower thanks to the KV cache. Specifically,
with saved KV cache from \prefilling, all the previous tokens do not need to pass any linear layers in the model.

\subsection{Inference with \methodname}
\label{sec:lazyllm_inference}
The overview of the proposed \methodname framework is illustrated in \cref{fig:overview}. \methodname starts with the full context and progressively prunes tokens to gradually reduce the number of computations towards the end of the model. Note, \methodname allows the model to select different subsets of tokens from the context in different generation steps,
even though some of them may be pruned in previous steps. Compared to static pruning which prunes all the tokens at once, dynamic pruning optimizes the next token prediction in each generation step, which is crucial to retaining the performance.

\textbf{Progressive Token Pruning.} 
Prior to this work, token pruning has been successfully applied to optimize LLM inference \citep{zhang2024h2o, li2024snapkv, adnan2024keyformer, nawrot2024dynamic}. However, these approaches require accumulating the full attention maps of predicting the first few tokens to profile the importance of prompt tokens before starting pruning. Consequently, they are not applicable to reduce \ttft as they still require computing all the KV cache at the \prefilling stage. 

In contrast,  \methodname only ``lazily'' computes the tokens that are important to predict the next token by starting from the \textit{first} iteration of the inference (the \prefilling step). A key challenge to pruning tokens in the first iteration is determining their importance. 
Inspired by the early exiting work \citep{elhoushi2024layer} which shows the token hidden states gradually evolve through the transformer layers, we apply layer-wise token pruning in each generation step.
Specifically, we use the attention map of the layer $A^{l}\in\mathcal{R}^{H\times N\times N}$ to determine the importance of input token $t_i$ \wrt the next token to be predicted as 
\begin{equation}
    s_i^l = \frac{1}{H} \sum_{h=1}^{H} A^{l}_{h, i, N}
\end{equation}
where $H$ denotes number of attention heads, $N$ is the sequence length, and $A_{h, i, j}$ is the attention probability of the token $t_j$ attending to token $t_i$ at $h^{th}$ head.

After computing the confidence scores of tokens, it is challenging to determine the threshold value to prune the token. Concretely, the threshold can change as the distribution of the attention scores varies between different layers and different tasks. We address this challenge by using the top-$k$ percentile selection strategy to prune tokens. Specifically, token $t_i$ is pruned at layer $l+1$ if its confidence score $s^l_i$ is smaller than $k^l$th percentile among the input tokens. Once the token is pruned, it is excluded from the computation of all successive layers. In other words, the tokens used in the later layers will be a subset of previous layers.

Our study in \cref{prune_ablation} shows the performance changes with different locations of pruning layers and the number of tokens pruned. In particular, when pruning at the same transformer layer, the model’s performance gradually decreases as fewer
tokens are kept. We also found pruning at later transformer layers consistently has better performance than pruning at earlier layers, suggesting that later layers are less sensitive to token pruning. To achieve a better balance of speedup and accuracy, as shown in \cref{fig:overview}, we apply progressive pruning that keeps more tokens at earlier transformer layers and gradually reduces the number of tokens towards the end of the transformer.

\input{figure_latex/method}

\textbf{Aux Cache.} In the prefilling stage, there is no KV cache and every token is represented by hidden states. Thus, progressive token pruning can be implemented by removing pruned tokens' hidden states.  However, extending the progressive token pruning to the following \decoding steps is non-trivial. This is because each \decoding step leverages the KV cache computed in the \prefilling to compute attention. As the \methodname performs progressive token pruning at the \prefilling stage, the KV of tokens pruned at layer $l$ (\eg $T4$ in \cref{fig:overview}) will not exist in the KV cache of layer $l+1$. As a reminder, the \methodname framework allows each generation step to pick a different subset set of tokens from the full input token sequences in every step, regardless of whether they are pruned in previous generation steps or not. For example, during the following \decoding steps, those pruned tokens (\eg $T4$) that do not exist in the KV cache of layer $l+1$ may be re-selected to compute attention. In such cases, the model can not retrieve the KV cache of these tokens. An intuitive solution is to pass those tokens again from the beginning of the transformer. However,
that would cause repetitive computation for the same token, and eventually slow down the whole
generation. 

To tackle this challenge, we introduce \emph{Aux Cache} in addition to the original KV cache, which stores the hidden states of those pruned tokens (\eg $T4$ and $T7$ in \cref{fig:overview}) if their KV is not present in the following layer's KV cache, which could be potentially retrieved for the following iterations. As shown in \cref{fig:overview}, in each \decoding step, each transformer layer (\eg layer $l+1$) first retrieves the KV cache of past tokens if they exist (\eg $T1$ and $T8$). For those tokens that do not exist in the KV cache (\eg $T3$), we could retrieve their hidden states from the \emph{Aux Cache} of its previous layer directly instead of passing through previous layers again. The introduction of \emph{Aux Cache} ensures that each token is computed at most once in every transformer layer, and ensures the worst runtime of \methodname is not slower than the baseline.

\section{Implementations Details} \label{apdx:imp_details}
We implement \methodname on Llama 2 \citep{touvron2023llama} and XGen \citep{nijkamp2023xgen} and evaluate it on the LongBench \citep{bai2023longbench} using HuggingFace\footnote{\url{https://github.com/huggingface/transformers/}}. We follow the official GitHub repository\footnote{\url{https://github.com/THUDM/LongBench}} of LongBench for data preprocessing and prompting in all experiments. The LongBench benchmark consists of multiple datasets in different tasks, where each task may have different metrics, including ROUGE-L, F1, Accuracy, and  Edit Sim. Following the official evaluation pipeline, we categorize all results over major task categories by computing the macro-average score. 

As previously noted, the proposed \methodname doesn't require any training. Thus, \methodname uses the exact same existing checkpoints as the baseline, for all models. For inference, we conduct all experiments on NVIDIA A100 GPUs. We measure and report the speedup based on the empirical walltime improvement. Specifically, for \emph{TTFT Speedup}, we measure the empirical walltime between when the prompt is fed to the model, and when the model generates the first token. For \emph{Generation Speedup}, we measure the empirical walltime between when the prompt is fed to the model, and when the model finished generating all output tokens. We add 5 warmup runs for each experiment before starting the time measurement to remove the noise such as loading model parameters.

%% file: figure_latex/teaser.tex
\begin{figure*}[t]
\centering
\includegraphics[width=\linewidth]{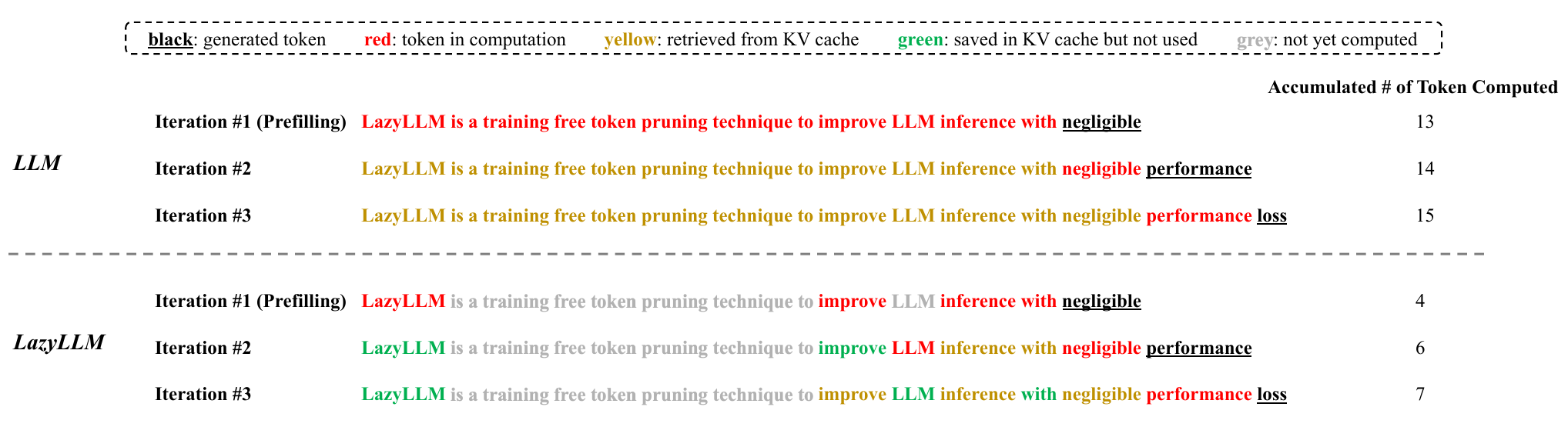}
\caption{Comparison between standard LLM and \methodname. Instead of computing the KV cache of all input tokens at the \prefilling stage, \methodname only selectively computes the tokens that are important to the next token prediction, deferring the computation of remaining tokens to later steps. \methodname significantly optimizes \ttft by reducing the amount of computation during \prefilling. Moreover, as some tokens in the prompt are never selected by \methodname during the whole generation process (even though theoretically the model \textit{could} use all tokens in the prompt),  \methodname also reduces the total amount of computation and accelerates the overall generation.}
\label{fig:teaser}
\end{figure*} 

%% file: figure_latex/method.tex
\begin{figure*}[t]
\centering
\includegraphics[width=\linewidth]{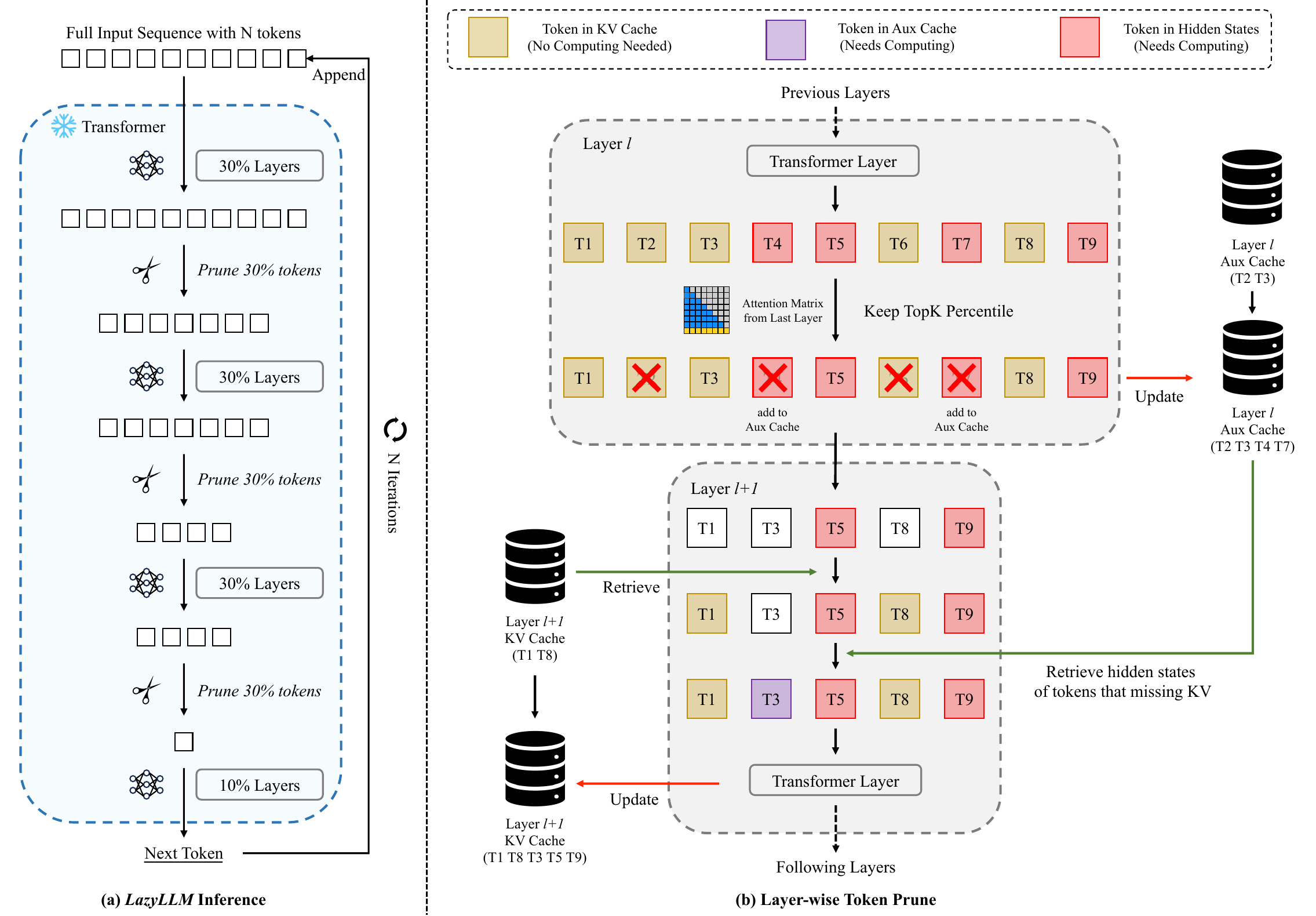}
\caption{Overview of the \methodname\ framework. \methodname starts with the full context and progressively prunes tokens to gradually reduce the number of computations towards the end of the model. \methodname allows the model to select different subsets of tokens from the context in different generation steps, which is crucial to retaining the performance.}
\label{fig:overview}
\end{figure*}

%% file: experiments.tex
\section{Experiments}

We examine our method using two large language models: Llama 2 7B and XGen 7B. We compare our method with baselines using the same publicly released pretrained checkpoints, without employing any additional training.  We perform experiments using LongBench, a multi-task benchmark for long content understanding. The LongBench comprises 16 datasets and covers 6 tasks including single-doc QA, multi-doc QA, summarization, few-shot learning, synthetic tasks, and code completion.

For the metrics, we primarily evaluate the effectiveness and efficiency of each method in the \ttft speedup \vs accuracy trade-off. Following LongBench, the accuracy (\emph{score}) denotes the macro-averaged scores across datasets in each task. The \ttft speedup measures the wall time improvement \wrt to the baseline for generating the first token. In analysis, we also assess the impact of our method on \emph{$\%$ of Prompt Token Computed} and \emph{Generation} speedup. The \emph{$\%$ of Prompt Token Computed} measures the accumulated percent of prompt tokens computed at the end of the generation, which indicates the save of total computation. The \emph{Generation} speedup measures the walltime change \wrt to the baseline for completing the entire generation process.

\input{tables/tasks_results}

\subsection{Results} \cref{tab:tasks_results} presents the \ttft speedup \vs accuracy comparisons between \methodname, standard LLM, and other baselines. In the table, the ``baseline'' refers to the standard LLM inference. The ``random token drop'' baseline is based on \citep{yao2022random} that randomly prunes the prompt tokens before feeding them to the LLMs. We report the average metrics across 5 runs for the ``random token drop'' baseline. Our ``static token pruning'' baseline prunes input tokens at once based on their attention score of the first few transformer layers during the \prefilling stage. We also compare with the prompt compression method \citep{li2023compressing} which pruning redundancy in the input context using LLMs. 
\cref{tab:tasks_results} shows \methodname consistently achieves better \ttft speedup with negligible accuracy drop across multiple tasks. It is worth noting that the overhead of running LLMs to compress the prompt is very computationally expensive. Even though the inference on the reduced prompt is faster, the actual \ttft of the ``prompt compression'' baseline is longer than the baseline.

\subsection{\ttft Speedup \vs Accuracy}

The inference efficiency of \methodname is controlled using three parameters: 1) the number of pruning layers, 2) the locations of these pruning layers, and 3) the number of tokens pruned within these layers. Increasing the number of pruning layers and pruning more tokens optimize computation by processing fewer tokens, and pruning tokens at earlier layers can save the computations for the successive layers. Prompting these factors will give more overall computation reduction, and offer better \ttft speedup. As a side effect, excessively pruning tokens may cause information loss and eventually lead to performance degradation. Similarly, the \ttft speedup and accuracy of baselines can vary with different hyperparameters.

We compare \ttft speedup \vs accuracy in \cref{fig:task_pf} with different hyperparameters. The visualization shows that, without any training, the proposed \methodname retains the accuracy better than baselines under the same \ttft speedup. For example, our method can offer $2.34\times$ \ttft speedup in the multi-document question-answering task with negligible ($\le 1\%$) performance loss. By controlling the pruning parameters, \methodname provides a good trade-off between accuracy and inference speed as compared to baseline methods. For instance, \methodname can achieve $3.0\times$ \ttft speedup in the multi-document question-answering task with $\le 10\%$ degradation in accuracy. On the other hand, baseline methods accuracy degrades significantly for similar \ttft speed-up. Note that the prompt compression approaches fail at improving \ttft because of the compression overhead.

\subsection{Impact on Overall Generation Speed}
To evaluate the impact of the proposed method on the overall generation process, we also profile the \emph{$\%$ of Prompt Token Computed} and \emph{Generation} speedup in \cref{tab:average_token}. We can find the \emph{$\%$ of Token Computed} of \methodname is less than 100$\%$, indicating that not all tokens in the prompt are selected by \methodname at the end of the generation, even though theoretically the model \textit{could} use all tokens. Computations in the FFN layers increase linearly, while those in the attention layers grow quadratically with the \emph{$\%$ of Token Computed}. A lower \emph{$\%$ of Token Computed} indicates \methodname reduces the total computation, consequently offering additional speedup to the overall generation process across diverse tasks.

\input{figure_latex/task_pf}

\input{tables/average_token}

\subsection{Drop Rate in Different Layers}
\label{prune_ablation}
In this section, we analyze the effect of the locations of pruning layers, and the number of tokens pruned. In particular, we report a series of experiments using a simplified version of \methodname that prunes tokens just once within the transformer. For each trial, we position the pruning layer at various levels of the transformer stack and apply different pruning ratios. We perform the experiments for both Llama 2 and XGen, and visualize the results in  \cref{fig:hp_ablation}. 

The results show both models share a similar trend. As expected, when pruning at the same transformer layer, the model's performance gradually decreases as fewer tokens are kept. Furthermore, pruning at later transformer layers consistently yields better performance compared to pruning at earlier layers, suggesting that later layers are less sensitive to token pruning. 
Based on these observations, we propose progressive token pruning in \cref{sec:lazyllm_inference}, which strategically prunes more tokens in later layers while preserving more in the earlier layers, optimizing the balance between efficiency and performance retention.

\input{figure_latex/hp_ablation}

\begin{figure}[hbt!]
\centering
\includegraphics[width=0.8\textwidth]{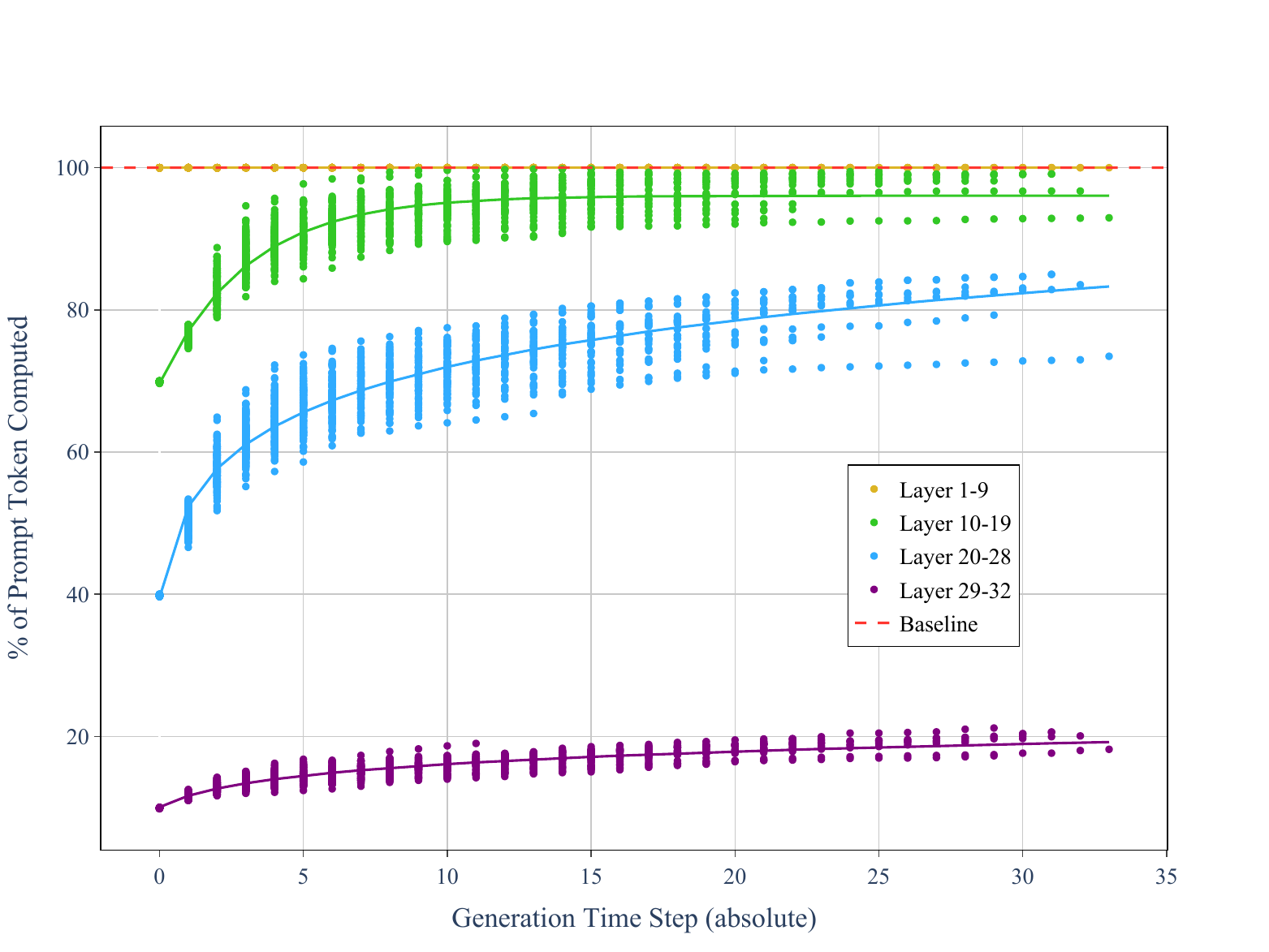}
\caption{
    Statistics on number of tokens processed during generation using our \textit{LazyLLM} technique with Llama 2 7B \citep{touvron2023llama}.
    We visualize the statistics of 1000 samples randomly sampled from LongBench.
    The $x$-axis represents the (absolute) generation time step, and the $y$-axis represents the number of prompt tokens processed at that time step (normalized by the prompt size).
    We visualize these statistics for various stages within the network.
    Note that cumulative token usage is upper-bounded by the baseline (evident with early layers).
}
\label{fig:layer_analysis}
\end{figure}

\subsection{Progressive KV Growth}

In this section, we characterize the internals of the model with the token pruning logic.
Specifically, we seek to understand what fractions of prompt tokens are cumulatively used and, inversely, not used.
This ``cumulative token usage'' can be equivalently defined as the KV cache size at each given step.
\cref{fig:layer_analysis} presents these cumulative prompt token usage numbers for each of the stages of the LazyLLM.

Our analysis supports the hypothesis that many tokens are never selected by the model (even though theoretically the model \textit{could} use all tokens in the prompt).
Since this model retains accuracy on the task(s), we can conclude that the model effectively drops the tokens which do not affect the output quality.

%% file: tables/tasks_results.tex
\begin{table*}[h!]
    \centering
    \resizebox{0.95\columnwidth}{!}{
    \begin{tabular}{lcccccc}
    \toprule[1.5pt]  
    \multirow{2}{*}{Tasks} & \multirow{2}{*}{Method} & \multicolumn{2}{c}{Llama 2} && \multicolumn{2}{c}{XGen} \\
    \cmidrule[1.25pt]{3-4}\cmidrule[1.25pt]{6-7}
    & & Score & TTFT Speedup ($\times$) && Score & TTFT Speedup ($\times$) \\
    \midrule[1.25pt]  
    \multirow{5}{*}{Single-Document QA} & Baseline & $\mathbf{25.79}$ & $1.00$ && $\mathbf{25.19}$ & $1.00$ \\
        & Random Token Drop & $20.05$ & $1.20$ && $18.32$ & $1.58$ \\
        & Static Token Pruning & $21.89$ & $1.18$ && $19.30$ & $1.61$ \\
        & Prompt Compression & $22.88$ & $0.12$ && $15.31$ & $0.20$ \\
        & \emph{LazyLLM (Ours)} & $25.59$ & $\mathbf{1.36}$ && $25.00$ & $\mathbf{1.96}$ \\ \hline
    \multirow{5}{*}{Multi-Document QA} & Baseline & $\mathbf{22.43}$ & $1.00$ && $\mathbf{20.71}$ & $1.00$ \\
        & Random Token Drop & $16.77$ & $1.19$ && $14.86$ & $1.37$ \\
        & Static Token Pruning & $19.93$ & $2.16$ && $17.23$ & $2.11$ \\
        & Prompt Compression & $8.42$ & $0.13$ && $11.56$ & $0.19$ \\
        & \emph{LazyLLM (Ours)} & $22.31$ & $\mathbf{2.34}$ && $20.68$ & $\mathbf{2.65}$ \\ \hline
    \multirow{5}{*}{Summarization} & Baseline & $24.65$ & $1.00$ && $\mathbf{24.85}$ & $1.00$ \\
        & Random Token Drop & $24.39$ & $1.39$ && $24.47$ & $1.70$ \\
        & Static Token Pruning & $24.59$ & $1.33$ && $24.46$ & $1.65$ \\
        & Prompt Compression & $25.16$ & $0.12$ && $24.57$ & $0.17$ \\
        & \emph{LazyLLM (Ours)} & $\mathbf{24.75}$ & $\mathbf{1.46}$ && $24.74$ & $\mathbf{1.91}$ \\ \hline
    \multirow{5}{*}{Few-shot Learning} & Baseline & $\mathbf{62.90}$ & $1.00$ && $\mathbf{56.40}$ & $1.00$ \\
        & Random Token Drop & $53.93$ & $1.19$ && $46.35$ & $1.62$ \\
        & Static Token Pruning & $56.54$ & $2.16$ && $51.93$ & $3.17$ \\
        & Prompt Compression & $24.18$ & $0.10$ && $23.72$ & $0.15$ \\
        & \emph{LazyLLM (Ours)} & $62.81$ & $\mathbf{2.19}$ && $56.12$ & $\mathbf{3.42}$ \\ \hline
    \multirow{5}{*}{Synthetic} & Baseline & $4.97$ & $1.00$ && $5.40$ & $1.00$ \\
        & Random Token Drop & $3.57$ & $1.18$ && $2.53$ & $1.13$ \\
        & Static Token Pruning & $2.81$ & $2.15$ && $3.00$ & $4.14$ \\
        & Prompt Compression & $3.20$ & $0.12$ && $1.42$ & $0.17$ \\
        & \emph{LazyLLM (Ours)} & $\mathbf{4.98}$ & $\mathbf{2.89}$ && $\mathbf{5.66}$ & $\mathbf{4.77}$ \\ \hline
    \multirow{5}{*}{Code Completion} & Baseline & $\mathbf{55.18}$ & $1.00$ && $\mathbf{36.49}$ & $1.00$ \\
        & Random Token Drop & $44.92$ & $1.23$ && $32.34$ & $1.57$ \\
        & Static Token Pruning & $37.51$ & $1.84$ && $32.27$ & $2.97$ \\
        & Prompt Compression & $17.45$ & $0.49$ && $11.38$ & $0.69$ \\
        & \emph{LazyLLM (Ours)} & $53.30$ &  $\mathbf{1.94}$ && $36.47$ & $\mathbf{3.47}$ \\ 
    \bottomrule[1.5pt]
    \end{tabular}
    }
    \caption{Comparisons of \ttft speedup \vs accuracy on various tasks. Without requiring any training/finetuning, \methodname consistently achieves better \ttft speedup with negligible accuracy drop.
    Note that the prompt compression approach fails at improving \ttft because the overhead of running LLMs to compress the prompt is very computationally expensive.
    }
    \label{tab:tasks_results}
\end{table*}

%% file: figure_latex/task_pf.tex
\begin{figure*}[t]
\centering
\includegraphics[width=\linewidth]{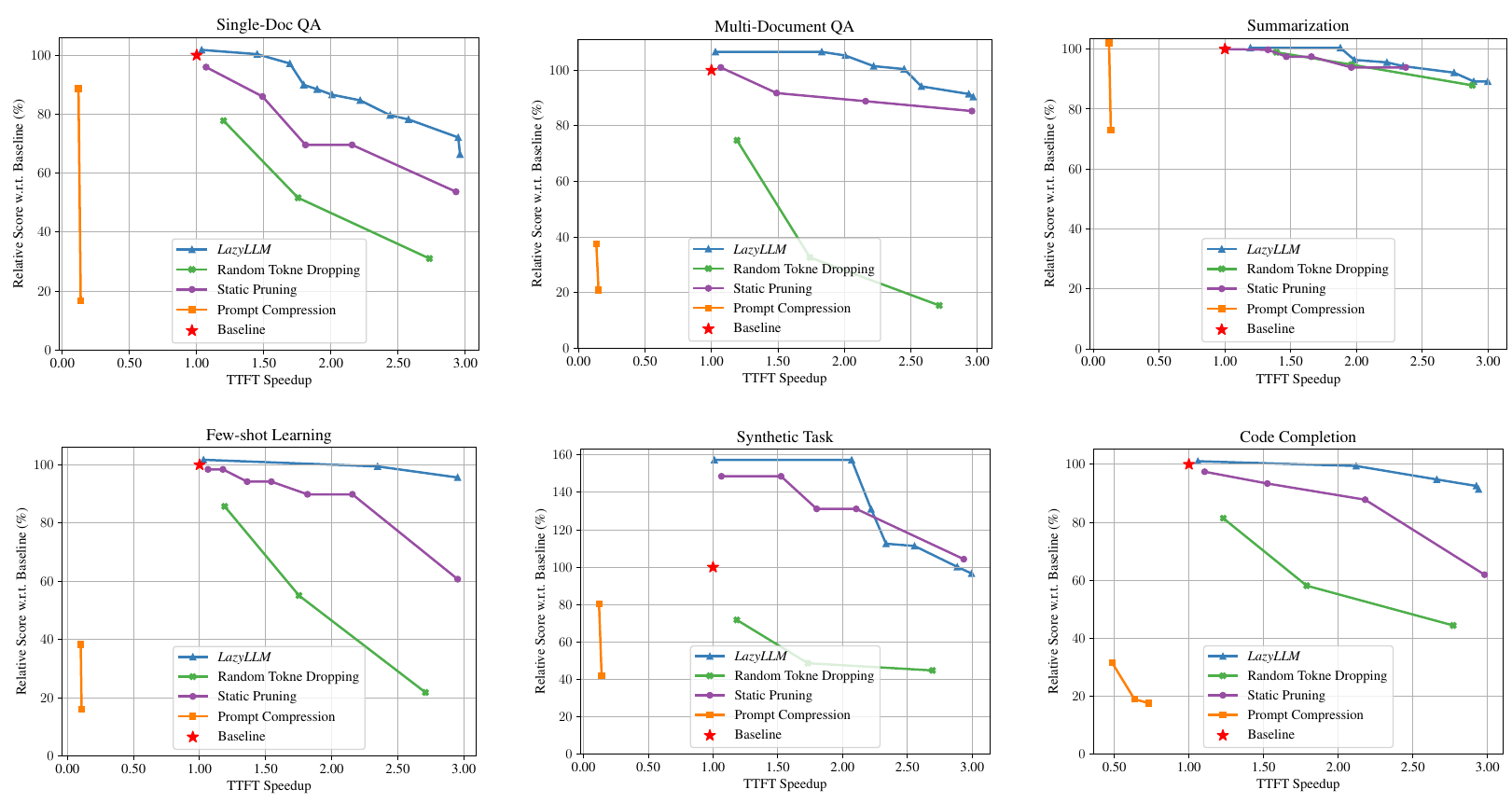}
\caption{\ttft speedup \vs accuracy comparison for Llama 2 7B across different tasks.}
\label{fig:task_pf}
\end{figure*}

%% file: tables/average_token.tex
\begin{table*}[hbt!]
    \begin{center}
    \begin{scriptsize}
    \begin{sc} 
    \centering
    \begin{tabular}{lccccc}
    \toprule[1.5pt]  
    \multirow{2}{*}{Tasks} & \multicolumn{2}{c}{$\%$ of Prompt Token Computed} && \multicolumn{2}{c}{Overall Generation Speedup} \\
    \cmidrule[1.25pt]{2-3}\cmidrule[1.25pt]{5-6}
     &  Llama 2 & XGen &&  Llama 2 & XGen \\
    \midrule[1.25pt]
     Single-Document QA & $87.31$ & $89.16$ && $1.34$ & $1.33$ \\
     Multi-Document QA & $63.94$ & $69.60$ && $1.56$ & $1.70$ \\
     Summarization & $99.59$ & $96.11$ && $1.02$ & $1.09$ \\
     Few-shot Learning & $69.98$ & $65.30$ && $1.28$ & $1.59$ \\
     Synthetic & $63.73$ & $40.54$ && $1.79$ & $3.16$ \\
     Code Completion & $68.57$ & $72.61$ && $1.01$ & $1.16$ \\
    \bottomrule[1.5pt]
    \end{tabular}
\end{sc}
\end{scriptsize}
    \caption{The \emph{$\%$ of Prompt Token Computed} and \emph{Generation} speedup of \methodname on various tasks. Reported values are based on the same setting as \cref{tab:tasks_results}. A lower \emph{$\%$ of Token Computed} indicates \methodname\ reduces the total computation, consequently offering additional speedup to the overall generation process across diverse tasks. }
    \label{tab:average_token}
\end{center}
\end{table*}

%% file: figure_latex/hp_ablation.tex
\begin{figure*}[t]
\centering
\includegraphics[width=\linewidth]{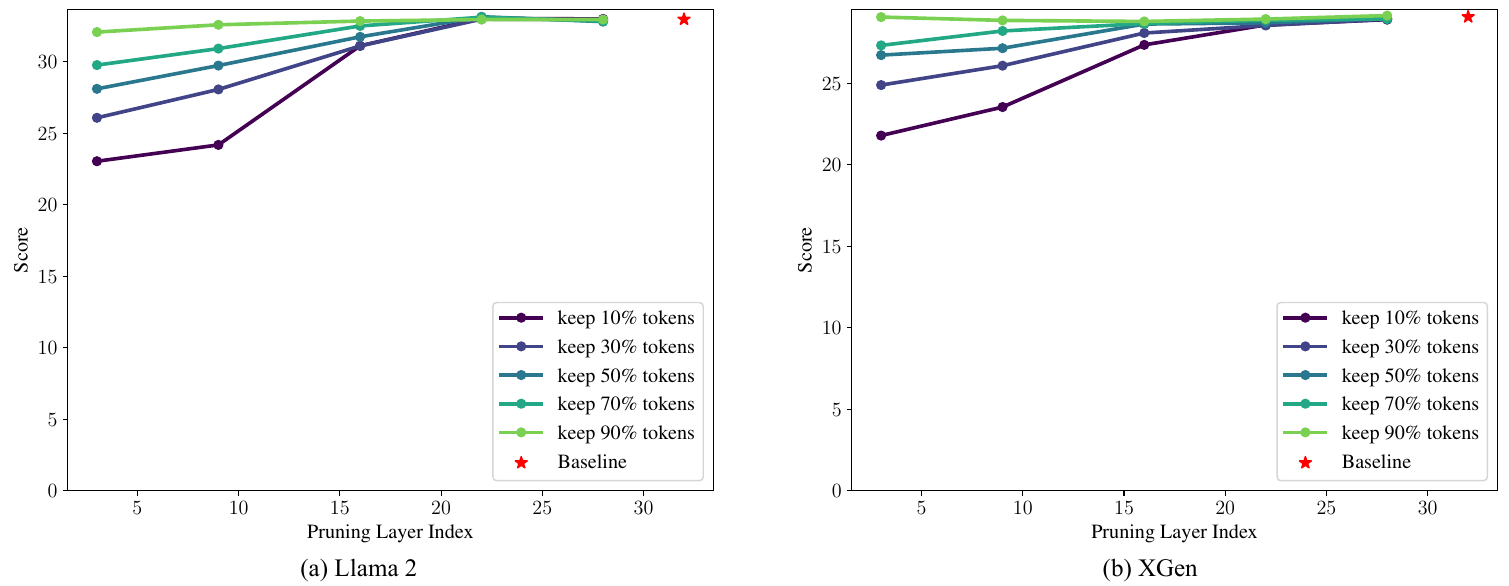}
\caption{Effect of the locations of pruning layers, and the number of tokens pruned. The results of both Llama 2 7B \cite{touvron2023llama} and XGen 7B \cite{nijkamp2023xgen} share a similar trend: 1) when pruning at the same transformer layer, the model's performance gradually decreases as fewer tokens are kept, and 2) Pruning at later transformer layers consistently has better performance than pruning at earlier layers, suggesting that later layers are less sensitive to token pruning. }
\label{fig:hp_ablation}
\end{figure*} 

%% file: conclusion.tex
\section{Conclusion}
In this work, we proposed a novel \methodname technique for efficient LLM inference, in particular under long context scenarios. \methodname selectively computes the KV for tokens important for the next token prediction and ``lazily'' defers the computation of remaining tokens to later steps, when they become relevant. We carefully examine \methodname on various tasks, where we observed the proposed method effectively reduces \ttft with negligible performance loss. 
It is worth noting that our method can be seamlessly integrated with existing transformer-based LLMs to improve their inference speed without requiring any fine-tuning. 